%% file: ms.tex
\documentclass[12pt, a4paper]{article}

\pdfoutput=1
 
\usepackage{parskip}
\usepackage[utf8]{inputenc} 
\usepackage[T1]{fontenc} 
\usepackage{hyperref} 
\usepackage{url} 
\usepackage{booktabs} 
\usepackage{amsfonts} 
\usepackage{nicefrac} 
\usepackage{microtype} 
\usepackage{xcolor} 

\usepackage[margin=1.20in]{geometry}
\usepackage{multirow}
\usepackage{graphicx}
\usepackage{amsmath}
\usepackage{amssymb}
\usepackage{hyperref}
\usepackage{csquotes}
\usepackage{caption}
\usepackage{tikz}

\usepackage[utf8]{inputenc}
\usepackage[british]{babel}
\usepackage[style=authoryear, maxcitenames=2, bibencoding=utf8, natbib=true,
 backend=biber]{biblatex}
\DeclareLanguageMapping{british}{british-apa}
\usepackage{url}
\usepackage[hyphenbreaks]{breakurl}

\usepackage[acronym]{glossaries}

\newacronym{fpt}{FPT}{facial processing technologies}
\newacronym{irb}{IRB}{Institutional Review Board}
\newacronym{eci}{ECI}{Election Commission of India}
\newacronym{ut}{UT}{Union Territories}
\newacronym{api}{API}{Application Programming Interface}

\title{Cinderella's shoe won't fit Soundarya: An audit of facial processing tools on Indian faces}

\author{Gaurav Jain\thanks{Fellow, Young Leaders in Tech Policy Fellowship} \space and \space Smriti Parsheera\thanks{Fellow, CyberBRICS Project and PhD Candidate, School of Public Policy, Indian Institute of Technology, Delhi.}\footnote{We thank Renuka Sane, Ajay Shah, Noopur Raval, Vidushi Marda, Robin Zachariah Tharakan, Rishab Bailey and Devendra Damle for valuable discussions and comments. Earlier drafts of this paper were discussed in meetings of the Data Governance Network, in a session on `AI Bias Beyond the Western Lens: Perspectives From India' organised by the authors at Mozilla Festival, 2021 and at the CVPR 2021 Workshop on `Beyond Fairness: Towards a Just, Equitable, and Accountable Computer Vision' organised by Timnit Gebru and Emily Denton. We are grateful to the participants at these events for inputs that helped us in improving the drafts. All errors are our own.}}

\date{\today}

\begin{document}

\maketitle

\begin{abstract}
The increasing adoption of facial processing systems in India is fraught with concerns of privacy, transparency, accountability, and missing procedural safeguards. At the same time, we also know very little about how these technologies perform on the diverse features, characteristics, and skin tones of India's 1.34 billion plus population. In this paper, we test the face detection and facial analysis functions of four commercial facial processing tools on a dataset of Indian faces. The tools display varying error rates in the face detection and gender and age classification functions. The gender classification error rate for Indian female faces is consistently higher compared to that of males -- the highest female error rate being 14.68\%. In some cases, this error rate is much higher than that shown by previous studies for females of other nationalities. Age classification errors are also high. Despite taking into account an acceptable error margin of plus or minus 10 years from a person's actual age, age prediction failures are in the range of 14.3\% to 42.2\%. These findings point to the limited accuracy of facial processing tools, particularly for certain demographic groups, and the need for more critical thinking before adopting such systems.

\end{abstract}

\newpage

\tableofcontents

\printglossary

\newpage

\input{01_intro}

\input{02_lit_review}

\input{03_methodology}
\input{04_ethical}

\input{05_analysis}
\input{06_conclusion}
\newpage
\printbibliography

\newpage


\end{document}

%% file: 01_intro.tex
\section{Introduction}

India is seeing a rise in the use of artificial intelligence (AI) based systems, with facial processing being one of the most common use cases. The AI Observatory contains a database of 65 automated decision-making systems in India, one-third of which relate to the use of facial recognition \autocite{joshi2020}. Project Panoptic, another useful tracker developed by the Internet Freedom Foundation, identifies 75 facial recognition projects \autocite{panoptic}. The applications of this technology, in the public and private sectors, include purposes such as law enforcement, employment screening, device security, attendance systems, and know your customer checks  \autocite{parsheera2019}. Amidst the COVID-19 crisis, government authorities also turned to testing facial authentication as one of the possible means of identity verification for accessing vaccines \autocite{barik2021}. 

India has a fairly rich body of literature documenting the concerns raised by the growing use of facial processing -- on grounds of lack of transparency, privacy concerns, biased outcomes, and a range of structural problems \autocites{vipra2021,bhandari2021,jain2020,aneja2020,kovacs2020,parsheera2019,marda2019,basu2019}. But empirical work on this subject still remains limited. This includes a dearth of studies auditing the performance of facial processing tools on the diversity of Indian faces, a gap that we aim to bridge through this paper. 

While focusing on accuracy of facial processing, specifically in terms of measurement of error rates, we would like to emphasise that the accuracy is a necessary but not sufficient condition for the adoption of this technology \autocite{parsheera2019}. Notably, even with 100 percent accuracy facial processing systems would still be fraught with concerns of privacy and accountability. In fact, as noted by \textcite{kalluri2020}, a more accurate facial recognition system only becomes a more potent weapon in the hands of exploitative companies and oppressive state agencies. Yet, the developers and adopters of \gls{fpt} continue to rely on claims of increased performance and accuracy to push for greater adoption of these systems. A critical evaluation of how well \gls{fpt} actually work, and for whom, therefore, becomes important. 

In this paper, we audit the face detection and analysis functions of four commercially available \gls{fpt} tools -- Microsoft Azure's Face, Amazon's Rekognition, Face++, and FaceX -- on a dataset of Indian faces. We do this using publicly available images of election candidates sourced from the website of the \gls{eci}. The goal is to understand how these tools perform in carrying out face detection, gender classification and age estimation functions on Indian faces, and the likely implications of errors in the results. 

Accuracy of facial processing is a highly context-specific metric. Its outcomes tend to vary based on the quality of images being used, the nature of the use case, and the characteristics of the demographic population. The diversity of Indian faces, consisting of over 1.34 billion individuals from a mix of racial, cultural, genetic, and environmental backgrounds, therefore, serves as a fertile ground for questioning the accuracy of facial processing systems. Given this diversity of the Indian population, it would be futile, and even erroneous, to try and classify `Indians’ as a distinct racial category \autocite{khan2021}. Accordingly, the references to `Indian faces' in this paper do not imply a racial categorisation but simply refer to the diversity of facial features exhibited by people living in the territory of India.

While espousing the need for research of this nature, we are also mindful of the sensitive nature of facial data and the ethical considerations that ought to guide this field of research. We, accordingly, focus on understanding the literature on ethical use of publicly available datasets \autocite{buchanan2021,vitaketal2016} and have suitably tried to apply the best practices on assessment of benefits and harms in the design and implementation of our work.

The rest of the paper is organised as follows. Section two presents a review of related work on auditing of \gls{fpt} tools and Indian face datasets. Section three outlines the methodology, including a description of our dataset and the selected methods of analysis. Section four discusses the ethical considerations that motivated the design of this study. Section five presents the findings of our research and their implications. We conclude in section six with a brief summary of the findings.

%% file: 02_lit_review.tex
\section{Related work}

We situate this work in the field of algorithmic auditing studies, particularly those that have focused on the performance of commercially available off-the-shelf facial processing tools \autocites{khalil2020,rajiBuolamwini2020,buolamwini2018_gendershades}. The Gender Shades study by Buolamwini and Gebru has been pioneering work in this field. Using a database of images of parliamentarians from  African and European countries, the authors reviewed the performance of the gender classification function of the \gls{fpt} tools offered by Microsoft, IBM and Face++.\footnote{The Pilot Parliaments Benchmark constructed by the authors for this purpose consisted of 1,270 images of parliamentarians from three African countries (Rwanda, Senegal, South Africa) and three European countries (Iceland, Finland, Sweden).} They found that the tools generated a disproportionately higher gender classification error, as high as 34.7\%, for darker-skinned females as compared to just 0.8\% for lighter-skinned males \autocite{buolamwini2018_gendershades}. Subsequent studies have extended a similar methodology to other tools like Amazon, Kairos, and Clarifai \autocites{rajiBuolamwini2019,scheuerman2019}.

\textcite{jung2018} also use a similar approach to infer the accuracy of gender, race, and age attribution features using multiple publicly available image datasets. But they do not relate the errors to variation in gender or skin tones. The authors of this study found that the tools offered by Microsoft, IBM, Face++ and Amazon were `generally proficient at determining gender' but all the tools performed poorly on inferring age \cite{jung2018}. A collective reading of these studies points to the importance of a granular exploration of the intersectionality of factors such as race, gender and age while studying AI bias.

The studies referred to above deal with different aspects of \textit{facial processing}, which is broader in scope than \textit{facial recognition}. Facial recognition refers to a one-to-one verification or one-to-many identification by matching two or more images \autocite{nist2020_testimony}. But facial processing also includes the functions of face detection (identifying a face in an image) and facial analysis (inferring characteristics such as gender, ethnicity, and emotions) \autocite{rajiBuolamwini2020}. Although our paper focuses on facial processing functions other than recognition, studies have found that the accuracy of facial recognition also tends to vary across demographic groups \autocite{KlareJain2012,snow2018aclu,NIST2019}. Instances of poor performance of facial recognition have also come to light in real world applications of the technology in India. As per admissions made by government agencies before the Delhi High Court, use of facial recognition for identification of missing persons had accuracy rates as low as 2 percent \autocite{pti2018}. Further, the software often resulted in the matching of pictures of missing boys as girls \autocite{pti2019}. 

Some of the literature referred to above has been instrumental in highlighting the limitations and pitfalls of facial processing. But just as much of the development of \gls{fpt} is taking place in contexts far removed from the Indian one, the critical discourse on facial processing is also coloured by western datasets and institutional perspectives \autocites{marda_newDelhiPolicing,sambasivan2021}.  It has been argued that the conventional or western-centric model of algorithmic fairness can not be directly applied to an Indian context due to different axes of discrimination in India, which encompasses factors like caste, religion, ethnicity and class \cite{sambasivan2021}. Further, differences in legal context, societal perspectives, state capacity, and institutional structures are also critical to understanding the politics and governance of AI systems \autocite{marda2019gis,parsheera2021}. In line with the calls for more context specific research on facial processing, this paper examines how available commercial \gls{fpt} tools perform on Indian faces and the implications of inaccurate results for different demographic groups. 

To the best of our knowledge, there has not been any empirical work on issues of accuracy or fairness focusing on Indian faces. Existing studies on facial processing in India can broadly be divided into two streams. The first consists of technical studies that focus on the applications and capabilities of facial processing. This covers issues such as classification of faces based on regional affiliations -- North and South Indian \autocite{India_NS_paper} or North, East and South Indian \autocite{sarin2020}, identification of genetic disorders in children \autocite{Narayanan2019}, and detection of emotions \autocite{iSAFE}. The second stream of work consists of research papers, reports and other critical perspectives on the use of facial processing, highlighting associated risks, harms, and modes of regulation \autocites{bhandari2021,joshi2020,jain2020,aneja2020,kovacs2020,parsheera2019,marda2019,basu2019}. We locate this paper in this latter steam of work but distinguish it from the existing studies in terms of the use of quantitative methods of analysis.

One of the reasons why empirical research auditing the performance of facial processing in India remains limited could be due to the scarcity of appropriate image datasets. \tablename~\ref{indian_faces_db} outlines the key features of the few India-specific facial datasets that are publicly available. As the table shows, most of these datasets cover only a limited number of unique individuals. Further, covered subjects are often university students, volunteers from major cities \autocite{lakshmi2021chicago, IIITMface,happy2017_ISED} or movie celebrities \autocite{IMFDB}, which limits their scope in terms of age, regional diversity and urban-rural representation. In addition to these, some Indian faces also find a place in popular fair computer vision datasets but are generally grouped under the broader category of South Asian faces \autocite{khan2021}. Realising the limitations of existing Indian faces datasets for meeting our objectives, we created a new dataset of facial images -- the \gls{eci} Faces Dataset -- using the methodology described below.

\input{Tables/indian_faces_db}

%% file: Tables/indian_faces_db.tex
\arraystretch{1.5}
\begin{table}[ht]
\begin{center}

\caption{Existing databases of Indian faces}

\label{indian_faces_db}

\begin{tabular}{p{0.25\textwidth}cccp{0.35\textwidth}}
\\[-1.8ex]\hline
\hline
Dataset & Faces & Individual & Age Range & Remark\\
\hline
NEI-DB \autocites{NEI-db} & 59,850 & 630 & Not Recorded & \small Covering 5 North-eastern states(both tribal and non-tribal faces)\\
CFD-India\autocites{lakshmi2021chicago} & 142 & 142 & 18 - 50 years & \small Volunteers from Delhi, representing different regions of India\\
IIITM-G \autocites{IIITMface} & 1,928 & 107 & Not Recorded & \small Students and staff of the ABV IIITM-G\\
IMFDB \autocites{IMFDB} & 34,512 & 100 & Broad estimates & \small Screenshots from 103 movies (5 languages); age is estimated\\
IFAD \autocites{reecha2015_IFAD}& 3,296 & 55 & Not available & \small Screenshots of Indian Hindi movies, age not estimated \\
ISED \autocites{happy2017_ISED} & 428 & 50 & 18 - 22 years & \small Students from IIT Kharagpur, representing different regions of India\\
\hline
\end{tabular}
\end{center}

\end{table}

%% file: 03_methodology.tex
 \section{Methodology} \label{sec3}

We created a new dataset using publicly available facial images from the \gls{eci}'s candidate affidavit portal.\footnote{Available at \url{https://affidavit.eci.gov.in/candidate-affidavit}, accessed on 1 December 2020.} This section presents a description of the \gls{eci} Faces Dataset and the manner in which it was processed. We also describe the criteria used for the selection of the chosen \gls{fpt} tools and the methods of analysis that were deployed.

\subsection{ECI Faces Dataset}

The \gls{eci}'s candidate portal contains photographs and other information submitted by electoral candidates in their affidavits. We gathered 49,346 observations of candidates who filed nominations to contest in the Parliamentary and State Assembly elections (both general and bye elections) held between May 2019 to Nov 2020. The data cleaning process was done in the following steps. 
\begin{enumerate}
\item Duplicate entries with the same image were removed. 
\item All entries with missing data or data entry errors, such as the candidate's age being described as zero, were removed. 
\item A manual check was done for multiple images of the same person after filtering the results for matching candidate and father/husband's name. 
\item Images which couldn't be processed by any one of the tools were removed. This resulted in the removal of 15 images. Notably, none of these images presented an error across all the tools, indicating tool specific image processing limitations. 

\end{enumerate}

Following this process, we arrived at a dataset of 32,184 unique observations, consisting of a photograph and details like State, age and gender for each entry. We, retained all valid data entries even though the \gls{eci} may have subsequently rejected some of these nominations for other reasons. For instance, this would include nominations filed by candidates below 25 years of age, which is the minimum age to stand for elections in India.\footnote{Articles 84(b) and 173 (b), Constitution of India.}

These 32,184 observation cover all the constituencies of India, which means that persons from all States and ~\gls{ut} are represented in the dataset. The \gls{eci} allows persons to select their gender as male, female, or third gender. In line with a 2014 decision by the Indian Supreme Court, the term third gender includes persons belonging to the \textit{hijra} community and transgender persons \autocite{nalsa2014} The dataset consists of 3,524 female (10.95\%), 28,646 male (89.01\%) and 14 third gender (0.04\%) candidates. Their ages range from 18 -- 100 years (see \tablename~\ref{image_df_stat}). 

\input{Tables/image_df_stat}

The data composition is influenced by various structural factors, like gender, caste, class, and religion, that shape who files an election nomination in India. Further, some States also held assembly elections in the period of data collection, leading to additional representation in the dataset compared to States and \gls{ut} covered only in the 2019 Parliamentary election. But, despite these limitations, the dataset is unique in its coverage and size and, therefore, forms a suitable basis for this research.

Most of the images in the dataset adhere to the Indian government's face image quality requirements for e-governance applications \autocite{goi2010_face_standard}. The standards require a front-facing image showing the face and shoulders of the person, with uniform lighting against a plain white or off-white/blue background. The photographs on the \gls{eci}'s portal could be digital images uploaded directly by the candidate or scanned versions of the physical photograph submitted with the affidavit. The submitted information is also verified by a designated returning officer \autocite{eci2020_affidavit, eci2020_online}, ensuring  a certain level of manual validation check before the images are made available on the portal.

\subsection{Facial processing tools}

The selection of the commercial~\gls{fpt} tool was done taking into account the tools covered in previous audit studies and the availability of gender and age classification features in those tools. Accordingly, we identified three tools, Microsoft Azure's Face, Amazon's Rekognition, and Face++, as being the most suitable for our purposes. We also added one tool of Indian origin - FaceX. This was done keeping in mind the studies that indicate that algorithms from a particular region tend to have comparatively more accurate results for images of persons belonging from that region \autocites{NIST2019, toole2011handbook}.

\subsection{Method of analysis}

In line with previous auditing studies \autocite{jung2018,buolamwini2018_gendershades,rajiBuolamwini2019}, we focus on studying the errors in the ~\gls{api} responses generated by the tools. We assess the accuracy of the predicted results by comparing the generated results with the candidate's self-disclosed gender and age information, which is treated as the ground truth. Two of the selected tools --  Amazon and FaceX -- generated an age range and not a value. In these cases, the mean of the provided range limits was used as the predicted age. But since FaceX provides age predictions from a set of pre-determined age ranges, we recognise that this may lead to higher error rates compared to other tools that either provide an age value or a dynamic age range.

The tools also vary in the presentation of their responses on gender classification, Microsoft and Face++ only mention the predicted gender without providing a probability for the prediction. On the other hand, Amazon and FaceX  provide a probability score to indicate how sure the model is of the predicted gender. To enable comparison across tools, we have not taken into account the probability scores in our analysis. Further, in line with the terminology used by the \gls{eci}'s affidavit portal and the explanation provided by the selected tools, we use the term gender, as opposed to sex, while discussing the findings and implications. This is also consistent with the individual's ability to self identify their gender while filing the election papers. 

As noted in the previous section, there are certain other grounds of marginalisation, like a person’s caste and tribal status, that are critical to studying fairness in the Indian context. However, this information is not available in the candidate information provided on the \gls{eci}'s affidavit portal.  While there are example of studies that have inferred caste details based on the last names of individuals \autocites{fisman2017_caste_loan,bhagavatula2019}, we find that these linkages are prone to errors. For instance, a given last name could correspond to multiple castes or tribes with variations based on the person's religion or geographic region. We also discarded the option of classifying the images based on the constituencies reserved for members of scheduled castes and tribes in each election as that would present an incomplete list of persons belonging to those groups. Our findings, therefore, do not include observations on caste or tribe based categorisation of Indian faces although the implications of incorrect attribute classification are likely to be all the more stark for persons belonging to these marginalised groups.

%% file: Tables/image_df_stat.tex
\begin{table}[ht]
\centering

\caption{Descriptive statistics about the dataset} 
\label{image_df_stat}

\begin{tabular}{lccc}
  \hline
Variable & N & Unique N & Percentage \\ 
  \hline
State/UT & 32184 & 36 & - \\ 
Gender &  &  &  \\ 
  - Female & 3524 & - & 10.95\% \\ 
  - Male & 28646 & - & 89.01\% \\ 
  - Third Gender & 14 & - & 0.04\% \\ 
 Age Brackets &  &  &  \\ 
  -  18 - 25 years & 55 &  & 0.17\% \\ 
  -  25 - 40 years & 10709 &  & 33.27\% \\ 
  -  40 - 55 years & 13936 &  & 43.3\% \\ 
  -  55 - 70 years & 6636 &  & 20.62\% \\ 
  -  70 - 85 years & 833 &  & 2.59\% \\ 
  -  85 - 100 years & 15 &  & 0.05\% \\ 
   \hline
\end{tabular}

\end{table}

%% file: 04_ethical.tex
\section{Ethical considerations} \label{sec4}

Given the sensitivity of facial biometric data and its intrusive applications, research on automated facial analysis raises several ethical concerns. Examples of problematic research studies include the use of ethnicity recognition tools to identify Chinese Uyghur Muslims \autocite{wang2018} and predicting sexual orientation using images from dating and social media sites \autocite{wangKosinski2018}. Such studies have been criticised for enabling the surveillance and prosecution of vulnerable groups \autocite{noorden2020,moreau2019} and violating expectations of privacy and sensitivity of the content \autocite{sweeney2017}. Besides being ethically problematic, researchers have highlighted that several use cases of facial processing, such as for emotion recognition, also rest on unsound technical and scientific foundations \autocite{vidushi2021}.  

While they may not generate the same types of concerns as applied AI research, algorithmic auditing studies are also bound by various ethical considerations. These considerations revolve around the source of the image dataset being used for the audit, its representativeness, and the rights of the covered individuals. Another concern that has been highlighted is that such studies may overemphasise the value of demographic representation in datasets rather than questioning the political or institutional context in which the technology is being used \autocite{hoffmann2019fair}.

Accordingly, we find it important to highlight some of the key ethical concerns that have informed this research. But, before doing so, we must emphasise that the scope of this paper is focused on testing the detection and classification of publicly available images by commercial \gls{fpt} tools. We have not used the available images for developing or training any algorithms or applied the tools for any particular use case. The methodology and objective of our work is therefore very different from the problematic types of applied AI studies referred to above.

The first issue that we had to address while designing this study related to that of individual consent. Obtaining consent is generally not regarded as necessary for research relying on publicly available data \autocite{buchanan2021}. India's Personal Data Protection Bill, 2019, which is still in a draft stage, also recognises the `processing of publicly available personal data' to be a reasonable purpose that can be pursued without consent from the individual.\footnote{Section 14(2)(g), Personal Data Protection Bill, 2019.} This is in addition to the proposed exemption for research purposes.\footnote{Section 38, Personal Data Protection Bill, 2019.}  Such an approach has, however, been questioned by some, particularly in the context of `public' data gleaned from social media sites \autocite{zimmer2010,sweeney2017,ravn2020}. Relying on Helen Nissenbaum’s conception of `contextual integrity', \textcite{vitaketal2016} note that while obtaining informed consent may not always be possible for online datasets, researchers should respect `the norms of the contexts in which online data was generated'.

The data on nomination affidavits filed by electoral candidates, which forms the basis of this research, is put out by the \gls{eci} for public information and transparency purposes. The \gls{eci}'s affidavit portal does not contain any restrictions on the download or use of its data. On the contrary, the \gls{eci} has clarified, the ``Affidavits and counter-affidavits are available for citizens to view and download" \autocite{eci2019}. Further, it can be presumed that the concerned persons have taken a conscious decision to be part of public life and put out their information for that purpose. While that does not imply an approval for being part of a facial processing audit, mandatory data put out by public figures arguably stands on a different footing from user-generated data on social media sites, which has more often been the subject of data ethics concerns. Taking into account the context of the data and the safeguards described below, we chose to proceed with the research without seeking the consent of individuals whose images were covered in the dataset.

We adopted the following safeguards to minimise any direct harm to the individuals and prevent any other potential negative societal implications.

\begin{description}
\item First, following the principle of data minimisation, we limited the data collection exercise only to the fields that were necessary to give effect to our research methodology. Details like political party affiliations and financial disclosures that are contained in the candidate affidavits but were not relevant to our study were not captured in the dataset.
\item Second, we selected a period of study during which a General Election had been held in the country. This ensured the inclusion of election candidates from all parts of the country in the dataset hence enhancing the representativeness of the data. While this data is still not fully representative of the population of India, it certainly performs better on this front compared to other publicly available Indian facial datasets.
\item Third, we excluded the evaluation of caste or tribe affiliations as inference of these affiliations from other available pieces of information like names or constituencies was fraught with the risk of falsely indicating the presence or absence of bias against particular groups on account of methodological limitations.
\item Fourth, in order to protect the privacy of the individuals, we have chosen not to disclose any personally identifiable information in the paper. This includes abstaining from the use of any illustrative images while presenting the results.
\item Fifth, we decided not to put out the gathered image dataset in the public domain. While this information is already available to the public, our decision to not release the dataset is motivated by the desire to avoid aiding potentially undesirable secondary uses of the cleaned and collated dataset. We have, however, released the code used to get the ~\gls{api} responses from the selected tools and the results generated by the processing of the images.\footnote{The code can be accessed at \url{https://github.com/gauravrpjain/auditing-fpt-eci-faces-db}} This can be used for establishing the verifiability of the results.
\end{description}

In light of these safeguards, we believe that the benefits of this research, in terms of furthering the agenda of algorithmic accountability by generating evidence on the performance of \gls{fpt} tools on Indian faces, outweigh any potential harms. While our objective is to further the agenda of algorithmic accountability, it is possible that the results may also be used by entities trying to make a case in favour of facial processing on the ground that the technology seems to work `in most cases'. We aim to mitigate this possibility by highlighting the serious implications of failed face detection and misclassification in the next section of the paper.

%% file: 05_analysis.tex
\section{Findings and implications} \label{sec5}

This section presents the results of our audit of the performance of the selected tools in carrying out the functions of face detection, gender classification and age estimation. We also discuss the implications of the observed errors taking into account some of the scenarios in which such functions may be deployed. This includes the problematic use case of deployment by police and investigative agencies for law enforcement purposes. While companies like Microsoft and Amazon have currently adopted bans and moratoriums on the supply of facial recognition technology to police authorities, there are numerous others that continue to do so \autocite{greig2021}. The use case specific discussions below, therefore, remain relevant even though the policies and detection and classification error rates may vary for different tools. 

\subsection{Face detection}

Since each image in the \gls{eci} Face Database is a headshot, face detection is considered to be a success if the tool is able to detect one face in the image. Detection error ($D$) represents a failure in detecting a face.
\[ D = \begin{cases}
0, &\text{if } n_{faces} = 1 \\
1, &\text{if } n_{faces} \neq 1 \\
\end{cases}
\]
where $n_{faces}$ is the number of faces detected. \tablename~\ref{detection_error_rate} reports the mean detection error for each tool and  \tablename~\ref{detection_success} provides the gender break-up of the successful cases of face detection. Only the successful cases of face detection are used for further analysis of the gender classification and age prediction functions.

\input{./Tables/detection_error_rate}
\input{./Tables/detection_success}

The four tools differ significantly in their ability to detect faces. Amazon performed the best, with an error rate of 0.05\% while Microsoft performed the worst, reporting an error rate of 3.17\%. To contextualise this in terms of numbers, out of the 32,184 persons in the dataset, Microsoft was unable to accurately detect the faces of a little over a thousand persons. This figure was similarly high for FaceX, which failed to detect 865 of the faces. While significant in themselves, the implications of these figures become all the more relevant when we take into account a population size of over 1.34 billion people, with many individuals being forced to encounter multiple facial processing projects.

The detection of a face in an image or live video is the logical starting point in any facial analysis process. Face detection could sometimes serve as an end in itself, for instance in the auto-focus feature of a camera, object detection by autonomous vehicles, or while estimating the size of a crowd. In most other cases, detection is followed by other types of processing, such as classification, emotion detection, identification or verification. Failure to have one's face detected, therefore, implies the system's inability to carry out any further processing, hence leading to the potential exclusion of the individual.

This may be regarded as a favourable outcome in some cases, like exclusion from detection by surveillance cameras. However, the ubiquitous adoption of facial recognition in multiple contexts means that non-detection could also impose significant inconvenience and costs. This can be particularly troubling when face recognition is used for purposes like access control, conducting know your customer checks, entry into airports and other buildings, delivery of welfare schemes and vaccine delivery \autocite{manikandan2020_frt_dbt,chakrabarty2020_ekyc,chandna2021_frt_cowin,lalwani2019_frt_airport_access}. Many of these services may offer an alternate route of manual human verification but availing the same could be subject to significant inconvenience, delays, and costs. Experience from the Aadhaar project also shows that manual alternatives to biometric authentication failures often do not work in practice, which could lead to the exclusion of legitimate beneficiaries \autocite{khera2019epw}.

\subsection{Gender classification}
Gender classification is considered to be a success if the predicted gender is the same as the self-described gender. The binary gender treatment of all tools, which assumes gender to be physiologically based and immutable \autocite{keyes2018,scheuerman2019}, makes the tools completely incapable of accurately classifying the gender of the fourteen persons belonging to the third gender in the dataset.\footnote{As per the documentation provided by Microsoft about its Azure tool, the possible values for gender are male, female, and genderless. However, there has been no cases of a face being classified as genderless in literature as well as in our dataset were classified as genderless.} 

For the remaining cases (where the self-described gender is either male or female) gender classification error ($G$) represents a failure in classifying gender.
\[ G = \begin{cases}
0, &\text{if } \hat{g} = g; n_{faces} = 1 \\
1, &\text{if } \hat{g} \neq g; n_{faces} = 1 \\
\end{cases}
\]
where $ \hat{g}\text{ is predicted gender; } \hat{g} \in \{F,M\}\text{ and }g\text{ is self described gender; } g \in \{F,M\}\text{.}$

\input{./Tables/gender_error_rate}

\tablename~\ref{gender_error_rate} reports the mean gender classification error rate for each tool. It shows that all the tools consistently fare poorer in the classification of female faces. This is particularly true in the case of Face++, which has a near-zero error rate for males but classifies 14.68\% of females as males. Face X and Amazon also reflect a noticeable gender error gap (difference between error for females and males) of 9.12\% and 2.14\%, respectively. These gender error gaps are statistically significant for all the tools. Figure 1 shows how the gender classification error varies across age for different gender groups. It reflects that the  chances of misclassification are consistently higher for females compared to males across all age groups. 

\begin{figure}
\begin{center}
\input{./IMG/gender_plot_age_error}%
\end{center}
\label{gender_plot_age_error}
\caption{Gender classification error rates across age groups}
\end{figure}

Of the selected tools, Amazon, Face++ and Microsoft have previously been covered by other gender classification studies \autocites{buolamwini2018_gendershades,rajiBuolamwini2019}. A comparison of our results with previous studies presents some interesting findings, particularly in the case of the Chinese company, Face++ (\tablename~\ref{facepp_gender_performance}). Our study finds error rates for Face++ to be as high as 14.68\% compared to an error rate of 2.5\% for females in the study by \textcite{rajiBuolamwini2019}. The error figures are more comparable for Amazon and Microsoft. This could possibly be an indication of the differences in the representation of Indian faces in the training dataset used by different firms and the relative importance of the Indian market for them. Further, the improved performance of Face++ between the 2018 and 2019 studies, which are based on the same data set (the Pilot Parliaments Benchmark) points to the tendency to introduce band-aid measures in response to specific studies while leaving the door open for other problems and inaccuracies.

\begin{table}[!htbp]
\begin{center}
\caption{Gender classification error rates for Face++ across studies}
\label{facepp_gender_performance}
\begin{tabular}{lccc}
\\[-1.8ex]\hline
Study & F & M & F-M \\
\hline
Buolamwini and Gebru (2018) & 21.3\% & 0.7\% & 20.6\% \\
Raji and Buolamwini (2019) & 2.5\% & 0.9\% & 1.6\% \\
Our study (2021) & 14.68\% & 0.26\% & 14.42\%\\
\hline
\end{tabular}\\
\end{center}
\end{table}

Some examples of automated gender classification include access control for gender-demarcated spaces and targeted advertisements \autocites{keyes2018,leufer2021}. Gender could also become the basis for automatic filtering of results for law enforcement purposes, such as identification of a suspect or a missing person based on their gender. Misclassification in such cases could result in incorrect identification, search or tracking of individuals. Similarly, use of gender analysis tools while filtering job applications could lead to unfair exclusions. The presence of humans in decision making chains could perhaps minimise some of these harms, but there is limited transparency and accountability about the oversight mechanisms incorporated by different users of such systems.

Finally, the use of facial processing could also have an impact on other models that do not explicitly include gender as a parameter for decision making. Research has shown a gendered bias to be present in all sorts of algorithmic systems from recruitment to text embeddings and even hate speech detectors \autocites{xia2020demoting,de2019bias,prates2019assessing,,sap2019risk,dixon2018measuring,parsheera2018}. Many of these studies illustrate how algorithms can acquire a gender bias even when gender is not an explicit input. Any model built using \glspl{fpt} for purposes beyond facial analysis could also implicitly extract gender information and contribute toward amplifying existing biases.

\subsection{Age prediction}

Recognising the challenges of estimating a person's exact age, we relied on age bins or brackets to calculate the age classification errors. Age prediction is considered a success if the difference between the actual age and the person's self-described age falls within the threshold limits, of 2, 5 and 10 years. This creates acceptable age bins of 5, 11 and 21 years, respectively. Age prediction error ($A$) represents a failure to predict age successfully.
\[ A = \begin{cases}
0, &\text{if } \mid\hat{a} - a\mid \leq t; n_{faces} = 1 \\
1, &\text{if } \mid\hat{a} - a\mid > t; n_{faces} = 1 \\
\end{cases}
\]
where $\hat{a}\text{ is predicted age, } a\text{ is self described age, and } t\text{ is threshold limit; } t \in \{2,5,10\}\text{.} $

\input{./Tables/age_error_rate}

\tablename~\ref{age_error_rate} reports the mean age error rate for each tool for the different threshold limits while Figure 2 highlights how the difference between the predicted age and self described age ($\hat{a} - a$) varies across age groups.

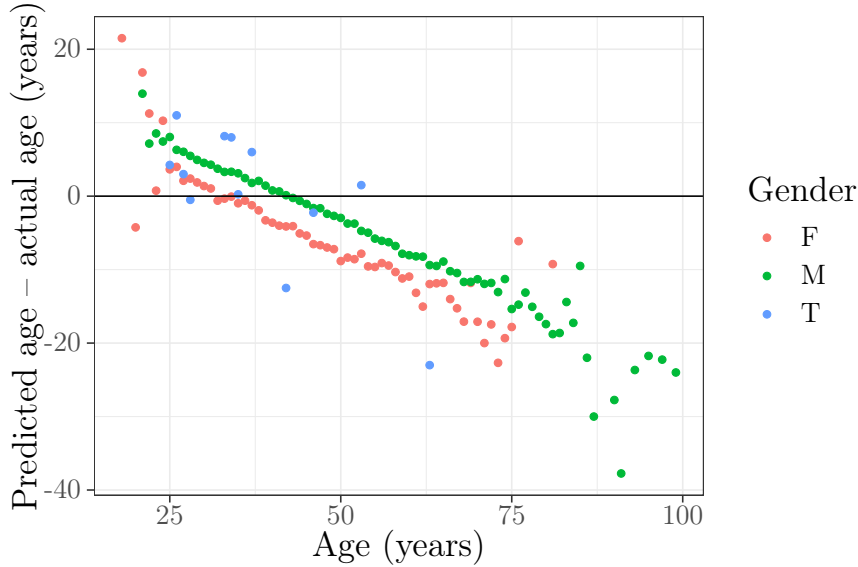
\begin{figure}
\begin{center}
\input{./IMG/age_plot_gender_error}%
\end{center}
\label{age_plot_gender_error}
\caption{Difference between predicted and self described age across gender groups}
\end{figure}

We find that all the tools perform fairly poorly on age prediction for a 5 year age bin, with error rates ranging from 71\% to 84\%. The error rates continue to remain high even when we significantly expand the size of the acceptable age range from 5 years to 21 years. FaceX performs the worst in this scenario with an error rate of 42.19\%. Further, Figure 2 highlights that tools tend to err in favour of predicting lower age ranges, which can make age prediction harder for older population groups.

Even though age prediction is generally recognised to be a hard to solve problem \autocite{jung2018}, suggested application areas of age estimation are not rare. For instance, age estimation may be used to monitor the use of age-appropriate online content \autocite{weiss2021} or in recruitment processes \autocite{angulu2018}. These processes may also form the basis for automatic filtering of results for law enforcement purposes, such as identification of a suspect or a missing person based on their gender or age. Misclassification in such cases could result in incorrect identification, search or tracking, causing significant harm to the affected individual. The poorer performance of the tools in the classification of vulnerable groups like the elderly only exacerbates these concerns.

%% file: Tables/detection_error_rate.tex
\begin{table}[ht]
\centering

\caption{Error Rates in face detection across FPT tools} 
\label{detection_error_rate}

\begin{tabular}{lccccc}
  \hline
Tool & N & \multicolumn{3}{c}{Faces Detected} & Overall \\ 
\cline{3-5}
  & & Zero & Multiple & One & \\
  \hline
Amazon & 32,184 & 7 & 9 & 32,168 & 0.05\% \\ 
Face++ & 32,184 & 586 & 2 & 31,596 & 1.83\% \\ 
FaceX & 32,184 & 649 & 213 & 31,319 & 2.69\% \\ 
Microsoft & 32,184 & 1,018 & 3 & 31,163 & 3.17\% \\ 
   \hline
\end{tabular}

\end{table}

%% file: Tables/detection_success.tex
\begin{table}[ht]
\centering

\caption{Successfully detected faces for facial analysis} 
\label{detection_success}

\begin{tabular}{lcccc}
\\[-1.8ex]\hline
\hline
Tool & No. & Female(F) & Male(M) & Third(T) \\
\hline
Amazon & 32,168 & 3,522 & 28,632 & 14 \\ 
Face++ & 31,596 & 3,480 & 28,102 & 14 \\ 
FaceX & 31,319 & 3,475 & 27,830 & 14 \\ 
Microsoft & 31,163 & 3,438 & 27,711 & 14 \\ 
\hline
\end{tabular}

\end{table}

%% file: Tables/gender_error_rate.tex
\begin{table*}[ht]
\centering

\caption{Error rates for gender classifcation across FPT tools} 
\label{gender_error_rate}

\begin{tabular}{lcccccc}
\\[-1.8ex]\hline
\hline
Tool & No. & Overall & Female(F) & Male(M) & F-M & t-test \\
\hline
Amazon & 32,154 & 0.66\% & 2.56\% & 0.42\% & 2.13\% & 7.94*** \\ 
Face++ & 31,582 & 1.85\% & 14.68\% & 0.26\% & 14.42\% & 24.01*** \\ 
FaceX & 31,305 & 2.36\% & 10.47\% & 1.35\% & 9.12\% & 17.41*** \\ 
Microsoft & 31,149 & 0.21\% & 0.79\% & 0.13\% & 0.65\% & 4.28*** \\ 
\hline
\textit{Note:} & \multicolumn{6}{r}{$^{.}$p$<$0.1;$^{*}$p$<$0.05; $^{**}$p$<$0.01; $^{***}$p$<$0.001} \\
\end{tabular}

\end{table*}

%% file: IMG/gender_plot_age_error.tex
\begin{tikzpicture}[x=1pt,y=1pt]
\definecolor{fillColor}{RGB}{255,255,255}
\path[use as bounding box,fill=fillColor,fill opacity=0.00] (0,0) rectangle (325.21,216.81);
\begin{scope}
\path[clip] (  0.00,  0.00) rectangle (325.21,216.81);
\definecolor{drawColor}{RGB}{255,255,255}
\definecolor{fillColor}{RGB}{255,255,255}

\path[draw=drawColor,line width= 0.6pt,line join=round,line cap=round,fill=fillColor] (  0.00,  0.00) rectangle (325.21,216.81);
\end{scope}
\begin{scope}
\path[clip] ( 31.71, 30.69) rectangle (262.78,211.31);
\definecolor{fillColor}{RGB}{255,255,255}

\path[fill=fillColor] ( 31.71, 30.69) rectangle (262.78,211.31);
\definecolor{drawColor}{gray}{0.92}

\path[draw=drawColor,line width= 0.3pt,line join=round] ( 31.71, 55.32) --
	(262.78, 55.32);

\path[draw=drawColor,line width= 0.3pt,line join=round] ( 31.71, 88.16) --
	(262.78, 88.16);

\path[draw=drawColor,line width= 0.3pt,line join=round] ( 31.71,121.00) --
	(262.78,121.00);

\path[draw=drawColor,line width= 0.3pt,line join=round] ( 31.71,153.84) --
	(262.78,153.84);

\path[draw=drawColor,line width= 0.3pt,line join=round] ( 31.71,186.68) --
	(262.78,186.68);

\path[draw=drawColor,line width= 0.3pt,line join=round] ( 92.79, 30.69) --
	( 92.79,211.31);

\path[draw=drawColor,line width= 0.3pt,line join=round] (157.62, 30.69) --
	(157.62,211.31);

\path[draw=drawColor,line width= 0.3pt,line join=round] (222.46, 30.69) --
	(222.46,211.31);

\path[draw=drawColor,line width= 0.6pt,line join=round] ( 31.71, 38.90) --
	(262.78, 38.90);

\path[draw=drawColor,line width= 0.6pt,line join=round] ( 31.71, 71.74) --
	(262.78, 71.74);

\path[draw=drawColor,line width= 0.6pt,line join=round] ( 31.71,104.58) --
	(262.78,104.58);

\path[draw=drawColor,line width= 0.6pt,line join=round] ( 31.71,137.42) --
	(262.78,137.42);

\path[draw=drawColor,line width= 0.6pt,line join=round] ( 31.71,170.26) --
	(262.78,170.26);

\path[draw=drawColor,line width= 0.6pt,line join=round] ( 31.71,203.10) --
	(262.78,203.10);

\path[draw=drawColor,line width= 0.6pt,line join=round] ( 60.37, 30.69) --
	( 60.37,211.31);

\path[draw=drawColor,line width= 0.6pt,line join=round] (125.20, 30.69) --
	(125.20,211.31);

\path[draw=drawColor,line width= 0.6pt,line join=round] (190.04, 30.69) --
	(190.04,211.31);

\path[draw=drawColor,line width= 0.6pt,line join=round] (254.87, 30.69) --
	(254.87,211.31);
\definecolor{drawColor}{RGB}{240,69,70}
\definecolor{fillColor}{RGB}{240,69,70}

\path[draw=drawColor,line width= 0.4pt,line join=round,line cap=round,fill=fillColor] ( 42.22,121.00) circle (  1.43);

\path[draw=drawColor,line width= 0.4pt,line join=round,line cap=round,fill=fillColor] ( 47.40, 38.90) circle (  1.43);

\path[draw=drawColor,line width= 0.4pt,line join=round,line cap=round,fill=fillColor] ( 50.00, 93.63) circle (  1.43);

\path[draw=drawColor,line width= 0.4pt,line join=round,line cap=round,fill=fillColor] ( 52.59,121.00) circle (  1.43);

\path[draw=drawColor,line width= 0.4pt,line join=round,line cap=round,fill=fillColor] ( 55.18,203.10) circle (  1.43);

\path[draw=drawColor,line width= 0.4pt,line join=round,line cap=round,fill=fillColor] ( 57.78, 38.90) circle (  1.43);

\path[draw=drawColor,line width= 0.4pt,line join=round,line cap=round,fill=fillColor] ( 60.37, 63.68) circle (  1.43);

\path[draw=drawColor,line width= 0.4pt,line join=round,line cap=round,fill=fillColor] ( 62.96, 51.66) circle (  1.43);

\path[draw=drawColor,line width= 0.4pt,line join=round,line cap=round,fill=fillColor] ( 65.56, 75.84) circle (  1.43);

\path[draw=drawColor,line width= 0.4pt,line join=round,line cap=round,fill=fillColor] ( 68.15, 60.37) circle (  1.43);

\path[draw=drawColor,line width= 0.4pt,line join=round,line cap=round,fill=fillColor] ( 70.74, 56.77) circle (  1.43);

\path[draw=drawColor,line width= 0.4pt,line join=round,line cap=round,fill=fillColor] ( 73.34, 61.96) circle (  1.43);

\path[draw=drawColor,line width= 0.4pt,line join=round,line cap=round,fill=fillColor] ( 75.93, 51.96) circle (  1.43);

\path[draw=drawColor,line width= 0.4pt,line join=round,line cap=round,fill=fillColor] ( 78.52, 62.70) circle (  1.43);

\path[draw=drawColor,line width= 0.4pt,line join=round,line cap=round,fill=fillColor] ( 81.12, 67.53) circle (  1.43);

\path[draw=drawColor,line width= 0.4pt,line join=round,line cap=round,fill=fillColor] ( 83.71, 69.93) circle (  1.43);

\path[draw=drawColor,line width= 0.4pt,line join=round,line cap=round,fill=fillColor] ( 86.30, 54.27) circle (  1.43);

\path[draw=drawColor,line width= 0.4pt,line join=round,line cap=round,fill=fillColor] ( 88.90, 62.71) circle (  1.43);

\path[draw=drawColor,line width= 0.4pt,line join=round,line cap=round,fill=fillColor] ( 91.49, 64.98) circle (  1.43);

\path[draw=drawColor,line width= 0.4pt,line join=round,line cap=round,fill=fillColor] ( 94.08, 54.99) circle (  1.43);

\path[draw=drawColor,line width= 0.4pt,line join=round,line cap=round,fill=fillColor] ( 96.68, 60.74) circle (  1.43);

\path[draw=drawColor,line width= 0.4pt,line join=round,line cap=round,fill=fillColor] ( 99.27, 55.65) circle (  1.43);

\path[draw=drawColor,line width= 0.4pt,line join=round,line cap=round,fill=fillColor] (101.86, 56.14) circle (  1.43);

\path[draw=drawColor,line width= 0.4pt,line join=round,line cap=round,fill=fillColor] (104.46, 59.09) circle (  1.43);

\path[draw=drawColor,line width= 0.4pt,line join=round,line cap=round,fill=fillColor] (107.05, 51.61) circle (  1.43);

\path[draw=drawColor,line width= 0.4pt,line join=round,line cap=round,fill=fillColor] (109.64, 59.04) circle (  1.43);

\path[draw=drawColor,line width= 0.4pt,line join=round,line cap=round,fill=fillColor] (112.24, 63.74) circle (  1.43);

\path[draw=drawColor,line width= 0.4pt,line join=round,line cap=round,fill=fillColor] (114.83, 60.54) circle (  1.43);

\path[draw=drawColor,line width= 0.4pt,line join=round,line cap=round,fill=fillColor] (117.42, 56.25) circle (  1.43);

\path[draw=drawColor,line width= 0.4pt,line join=round,line cap=round,fill=fillColor] (120.02, 70.21) circle (  1.43);

\path[draw=drawColor,line width= 0.4pt,line join=round,line cap=round,fill=fillColor] (122.61, 69.08) circle (  1.43);

\path[draw=drawColor,line width= 0.4pt,line join=round,line cap=round,fill=fillColor] (125.20, 70.09) circle (  1.43);

\path[draw=drawColor,line width= 0.4pt,line join=round,line cap=round,fill=fillColor] (127.80, 65.01) circle (  1.43);

\path[draw=drawColor,line width= 0.4pt,line join=round,line cap=round,fill=fillColor] (130.39, 62.87) circle (  1.43);

\path[draw=drawColor,line width= 0.4pt,line join=round,line cap=round,fill=fillColor] (132.98, 66.73) circle (  1.43);

\path[draw=drawColor,line width= 0.4pt,line join=round,line cap=round,fill=fillColor] (135.58, 60.58) circle (  1.43);

\path[draw=drawColor,line width= 0.4pt,line join=round,line cap=round,fill=fillColor] (138.17, 75.55) circle (  1.43);

\path[draw=drawColor,line width= 0.4pt,line join=round,line cap=round,fill=fillColor] (140.76, 55.88) circle (  1.43);

\path[draw=drawColor,line width= 0.4pt,line join=round,line cap=round,fill=fillColor] (143.36, 69.30) circle (  1.43);

\path[draw=drawColor,line width= 0.4pt,line join=round,line cap=round,fill=fillColor] (145.95, 65.17) circle (  1.43);

\path[draw=drawColor,line width= 0.4pt,line join=round,line cap=round,fill=fillColor] (148.54, 61.62) circle (  1.43);

\path[draw=drawColor,line width= 0.4pt,line join=round,line cap=round,fill=fillColor] (151.14, 81.12) circle (  1.43);

\path[draw=drawColor,line width= 0.4pt,line join=round,line cap=round,fill=fillColor] (153.73, 62.69) circle (  1.43);

\path[draw=drawColor,line width= 0.4pt,line join=round,line cap=round,fill=fillColor] (156.32, 56.49) circle (  1.43);

\path[draw=drawColor,line width= 0.4pt,line join=round,line cap=round,fill=fillColor] (158.92, 57.14) circle (  1.43);

\path[draw=drawColor,line width= 0.4pt,line join=round,line cap=round,fill=fillColor] (161.51, 75.39) circle (  1.43);

\path[draw=drawColor,line width= 0.4pt,line join=round,line cap=round,fill=fillColor] (164.10, 71.02) circle (  1.43);

\path[draw=drawColor,line width= 0.4pt,line join=round,line cap=round,fill=fillColor] (166.70,101.03) circle (  1.43);

\path[draw=drawColor,line width= 0.4pt,line join=round,line cap=round,fill=fillColor] (169.29, 56.18) circle (  1.43);

\path[draw=drawColor,line width= 0.4pt,line join=round,line cap=round,fill=fillColor] (171.88, 96.85) circle (  1.43);

\path[draw=drawColor,line width= 0.4pt,line join=round,line cap=round,fill=fillColor] (174.48, 59.42) circle (  1.43);

\path[draw=drawColor,line width= 0.4pt,line join=round,line cap=round,fill=fillColor] (177.07, 49.16) circle (  1.43);

\path[draw=drawColor,line width= 0.4pt,line join=round,line cap=round,fill=fillColor] (179.66, 79.95) circle (  1.43);

\path[draw=drawColor,line width= 0.4pt,line join=round,line cap=round,fill=fillColor] (182.26, 79.95) circle (  1.43);

\path[draw=drawColor,line width= 0.4pt,line join=round,line cap=round,fill=fillColor] (184.85,121.00) circle (  1.43);

\path[draw=drawColor,line width= 0.4pt,line join=round,line cap=round,fill=fillColor] (187.45, 38.90) circle (  1.43);

\path[draw=drawColor,line width= 0.4pt,line join=round,line cap=round,fill=fillColor] (190.04, 38.90) circle (  1.43);

\path[draw=drawColor,line width= 0.4pt,line join=round,line cap=round,fill=fillColor] (192.63, 79.95) circle (  1.43);

\path[draw=drawColor,line width= 0.4pt,line join=round,line cap=round,fill=fillColor] (205.60, 38.90) circle (  1.43);
\definecolor{drawColor}{RGB}{98,199,107}
\definecolor{fillColor}{RGB}{98,199,107}

\path[draw=drawColor,line width= 0.4pt,line join=round,line cap=round,fill=fillColor] ( 50.00, 38.90) circle (  1.43);

\path[draw=drawColor,line width= 0.4pt,line join=round,line cap=round,fill=fillColor] ( 52.59, 56.18) circle (  1.43);

\path[draw=drawColor,line width= 0.4pt,line join=round,line cap=round,fill=fillColor] ( 55.18, 38.90) circle (  1.43);

\path[draw=drawColor,line width= 0.4pt,line join=round,line cap=round,fill=fillColor] ( 57.78, 38.90) circle (  1.43);

\path[draw=drawColor,line width= 0.4pt,line join=round,line cap=round,fill=fillColor] ( 60.37, 39.40) circle (  1.43);

\path[draw=drawColor,line width= 0.4pt,line join=round,line cap=round,fill=fillColor] ( 62.96, 39.49) circle (  1.43);

\path[draw=drawColor,line width= 0.4pt,line join=round,line cap=round,fill=fillColor] ( 65.56, 40.99) circle (  1.43);

\path[draw=drawColor,line width= 0.4pt,line join=round,line cap=round,fill=fillColor] ( 68.15, 39.26) circle (  1.43);

\path[draw=drawColor,line width= 0.4pt,line join=round,line cap=round,fill=fillColor] ( 70.74, 41.04) circle (  1.43);

\path[draw=drawColor,line width= 0.4pt,line join=round,line cap=round,fill=fillColor] ( 73.34, 40.94) circle (  1.43);

\path[draw=drawColor,line width= 0.4pt,line join=round,line cap=round,fill=fillColor] ( 75.93, 39.75) circle (  1.43);

\path[draw=drawColor,line width= 0.4pt,line join=round,line cap=round,fill=fillColor] ( 78.52, 39.80) circle (  1.43);

\path[draw=drawColor,line width= 0.4pt,line join=round,line cap=round,fill=fillColor] ( 81.12, 39.94) circle (  1.43);

\path[draw=drawColor,line width= 0.4pt,line join=round,line cap=round,fill=fillColor] ( 83.71, 40.49) circle (  1.43);

\path[draw=drawColor,line width= 0.4pt,line join=round,line cap=round,fill=fillColor] ( 86.30, 39.64) circle (  1.43);

\path[draw=drawColor,line width= 0.4pt,line join=round,line cap=round,fill=fillColor] ( 88.90, 40.95) circle (  1.43);

\path[draw=drawColor,line width= 0.4pt,line join=round,line cap=round,fill=fillColor] ( 91.49, 40.52) circle (  1.43);

\path[draw=drawColor,line width= 0.4pt,line join=round,line cap=round,fill=fillColor] ( 94.08, 40.58) circle (  1.43);

\path[draw=drawColor,line width= 0.4pt,line join=round,line cap=round,fill=fillColor] ( 96.68, 39.99) circle (  1.43);

\path[draw=drawColor,line width= 0.4pt,line join=round,line cap=round,fill=fillColor] ( 99.27, 40.17) circle (  1.43);

\path[draw=drawColor,line width= 0.4pt,line join=round,line cap=round,fill=fillColor] (101.86, 39.55) circle (  1.43);

\path[draw=drawColor,line width= 0.4pt,line join=round,line cap=round,fill=fillColor] (104.46, 40.36) circle (  1.43);

\path[draw=drawColor,line width= 0.4pt,line join=round,line cap=round,fill=fillColor] (107.05, 39.77) circle (  1.43);

\path[draw=drawColor,line width= 0.4pt,line join=round,line cap=round,fill=fillColor] (109.64, 40.17) circle (  1.43);

\path[draw=drawColor,line width= 0.4pt,line join=round,line cap=round,fill=fillColor] (112.24, 39.72) circle (  1.43);

\path[draw=drawColor,line width= 0.4pt,line join=round,line cap=round,fill=fillColor] (114.83, 40.41) circle (  1.43);

\path[draw=drawColor,line width= 0.4pt,line join=round,line cap=round,fill=fillColor] (117.42, 40.18) circle (  1.43);

\path[draw=drawColor,line width= 0.4pt,line join=round,line cap=round,fill=fillColor] (120.02, 40.70) circle (  1.43);

\path[draw=drawColor,line width= 0.4pt,line join=round,line cap=round,fill=fillColor] (122.61, 40.26) circle (  1.43);

\path[draw=drawColor,line width= 0.4pt,line join=round,line cap=round,fill=fillColor] (125.20, 40.10) circle (  1.43);

\path[draw=drawColor,line width= 0.4pt,line join=round,line cap=round,fill=fillColor] (127.80, 40.50) circle (  1.43);

\path[draw=drawColor,line width= 0.4pt,line join=round,line cap=round,fill=fillColor] (130.39, 41.52) circle (  1.43);

\path[draw=drawColor,line width= 0.4pt,line join=round,line cap=round,fill=fillColor] (132.98, 40.45) circle (  1.43);

\path[draw=drawColor,line width= 0.4pt,line join=round,line cap=round,fill=fillColor] (135.58, 40.59) circle (  1.43);

\path[draw=drawColor,line width= 0.4pt,line join=round,line cap=round,fill=fillColor] (138.17, 42.46) circle (  1.43);

\path[draw=drawColor,line width= 0.4pt,line join=round,line cap=round,fill=fillColor] (140.76, 40.90) circle (  1.43);

\path[draw=drawColor,line width= 0.4pt,line join=round,line cap=round,fill=fillColor] (143.36, 40.67) circle (  1.43);

\path[draw=drawColor,line width= 0.4pt,line join=round,line cap=round,fill=fillColor] (145.95, 41.68) circle (  1.43);

\path[draw=drawColor,line width= 0.4pt,line join=round,line cap=round,fill=fillColor] (148.54, 42.31) circle (  1.43);

\path[draw=drawColor,line width= 0.4pt,line join=round,line cap=round,fill=fillColor] (151.14, 40.18) circle (  1.43);

\path[draw=drawColor,line width= 0.4pt,line join=round,line cap=round,fill=fillColor] (153.73, 40.41) circle (  1.43);

\path[draw=drawColor,line width= 0.4pt,line join=round,line cap=round,fill=fillColor] (156.32, 41.91) circle (  1.43);

\path[draw=drawColor,line width= 0.4pt,line join=round,line cap=round,fill=fillColor] (158.92, 41.15) circle (  1.43);

\path[draw=drawColor,line width= 0.4pt,line join=round,line cap=round,fill=fillColor] (161.51, 42.14) circle (  1.43);

\path[draw=drawColor,line width= 0.4pt,line join=round,line cap=round,fill=fillColor] (164.10, 42.39) circle (  1.43);

\path[draw=drawColor,line width= 0.4pt,line join=round,line cap=round,fill=fillColor] (166.70, 44.22) circle (  1.43);

\path[draw=drawColor,line width= 0.4pt,line join=round,line cap=round,fill=fillColor] (169.29, 43.97) circle (  1.43);

\path[draw=drawColor,line width= 0.4pt,line join=round,line cap=round,fill=fillColor] (171.88, 40.28) circle (  1.43);

\path[draw=drawColor,line width= 0.4pt,line join=round,line cap=round,fill=fillColor] (174.48, 42.62) circle (  1.43);

\path[draw=drawColor,line width= 0.4pt,line join=round,line cap=round,fill=fillColor] (177.07, 43.87) circle (  1.43);

\path[draw=drawColor,line width= 0.4pt,line join=round,line cap=round,fill=fillColor] (179.66, 41.09) circle (  1.43);

\path[draw=drawColor,line width= 0.4pt,line join=round,line cap=round,fill=fillColor] (182.26, 43.30) circle (  1.43);

\path[draw=drawColor,line width= 0.4pt,line join=round,line cap=round,fill=fillColor] (184.85, 40.73) circle (  1.43);

\path[draw=drawColor,line width= 0.4pt,line join=round,line cap=round,fill=fillColor] (187.45, 42.47) circle (  1.43);

\path[draw=drawColor,line width= 0.4pt,line join=round,line cap=round,fill=fillColor] (190.04, 46.04) circle (  1.43);

\path[draw=drawColor,line width= 0.4pt,line join=round,line cap=round,fill=fillColor] (192.63, 49.49) circle (  1.43);

\path[draw=drawColor,line width= 0.4pt,line join=round,line cap=round,fill=fillColor] (195.23, 51.62) circle (  1.43);

\path[draw=drawColor,line width= 0.4pt,line join=round,line cap=round,fill=fillColor] (197.82, 52.21) circle (  1.43);

\path[draw=drawColor,line width= 0.4pt,line join=round,line cap=round,fill=fillColor] (200.41, 53.60) circle (  1.43);

\path[draw=drawColor,line width= 0.4pt,line join=round,line cap=round,fill=fillColor] (203.01, 44.66) circle (  1.43);

\path[draw=drawColor,line width= 0.4pt,line join=round,line cap=round,fill=fillColor] (205.60, 38.90) circle (  1.43);

\path[draw=drawColor,line width= 0.4pt,line join=round,line cap=round,fill=fillColor] (208.19, 38.90) circle (  1.43);

\path[draw=drawColor,line width= 0.4pt,line join=round,line cap=round,fill=fillColor] (210.79, 38.90) circle (  1.43);

\path[draw=drawColor,line width= 0.4pt,line join=round,line cap=round,fill=fillColor] (213.38, 38.90) circle (  1.43);

\path[draw=drawColor,line width= 0.4pt,line join=round,line cap=round,fill=fillColor] (215.97, 38.90) circle (  1.43);

\path[draw=drawColor,line width= 0.4pt,line join=round,line cap=round,fill=fillColor] (218.57, 87.55) circle (  1.43);

\path[draw=drawColor,line width= 0.4pt,line join=round,line cap=round,fill=fillColor] (221.16, 38.90) circle (  1.43);

\path[draw=drawColor,line width= 0.4pt,line join=round,line cap=round,fill=fillColor] (228.94, 38.90) circle (  1.43);

\path[draw=drawColor,line width= 0.4pt,line join=round,line cap=round,fill=fillColor] (231.53, 38.90) circle (  1.43);

\path[draw=drawColor,line width= 0.4pt,line join=round,line cap=round,fill=fillColor] (236.72, 38.90) circle (  1.43);

\path[draw=drawColor,line width= 0.4pt,line join=round,line cap=round,fill=fillColor] (241.91, 38.90) circle (  1.43);

\path[draw=drawColor,line width= 0.4pt,line join=round,line cap=round,fill=fillColor] (247.09, 38.90) circle (  1.43);

\path[draw=drawColor,line width= 0.4pt,line join=round,line cap=round,fill=fillColor] (252.28, 38.90) circle (  1.43);
\definecolor{drawColor}{gray}{0.20}

\path[draw=drawColor,line width= 0.6pt,line join=round,line cap=round] ( 31.71, 30.69) rectangle (262.78,211.31);
\end{scope}
\begin{scope}
\path[clip] (  0.00,  0.00) rectangle (325.21,216.81);
\definecolor{drawColor}{gray}{0.30}

\node[text=drawColor,anchor=base east,inner sep=0pt, outer sep=0pt, scale=  0.88] at ( 26.76, 35.87) {0};

\node[text=drawColor,anchor=base east,inner sep=0pt, outer sep=0pt, scale=  0.88] at ( 26.76, 68.71) {10};

\node[text=drawColor,anchor=base east,inner sep=0pt, outer sep=0pt, scale=  0.88] at ( 26.76,101.55) {20};

\node[text=drawColor,anchor=base east,inner sep=0pt, outer sep=0pt, scale=  0.88] at ( 26.76,134.39) {30};

\node[text=drawColor,anchor=base east,inner sep=0pt, outer sep=0pt, scale=  0.88] at ( 26.76,167.23) {40};

\node[text=drawColor,anchor=base east,inner sep=0pt, outer sep=0pt, scale=  0.88] at ( 26.76,200.07) {50};
\end{scope}
\begin{scope}
\path[clip] (  0.00,  0.00) rectangle (325.21,216.81);
\definecolor{drawColor}{gray}{0.20}

\path[draw=drawColor,line width= 0.6pt,line join=round] ( 28.96, 38.90) --
	( 31.71, 38.90);

\path[draw=drawColor,line width= 0.6pt,line join=round] ( 28.96, 71.74) --
	( 31.71, 71.74);

\path[draw=drawColor,line width= 0.6pt,line join=round] ( 28.96,104.58) --
	( 31.71,104.58);

\path[draw=drawColor,line width= 0.6pt,line join=round] ( 28.96,137.42) --
	( 31.71,137.42);

\path[draw=drawColor,line width= 0.6pt,line join=round] ( 28.96,170.26) --
	( 31.71,170.26);

\path[draw=drawColor,line width= 0.6pt,line join=round] ( 28.96,203.10) --
	( 31.71,203.10);
\end{scope}
\begin{scope}
\path[clip] (  0.00,  0.00) rectangle (325.21,216.81);
\definecolor{drawColor}{gray}{0.20}

\path[draw=drawColor,line width= 0.6pt,line join=round] ( 60.37, 27.94) --
	( 60.37, 30.69);

\path[draw=drawColor,line width= 0.6pt,line join=round] (125.20, 27.94) --
	(125.20, 30.69);

\path[draw=drawColor,line width= 0.6pt,line join=round] (190.04, 27.94) --
	(190.04, 30.69);

\path[draw=drawColor,line width= 0.6pt,line join=round] (254.87, 27.94) --
	(254.87, 30.69);
\end{scope}
\begin{scope}
\path[clip] (  0.00,  0.00) rectangle (325.21,216.81);
\definecolor{drawColor}{gray}{0.30}

\node[text=drawColor,anchor=base,inner sep=0pt, outer sep=0pt, scale=  0.88] at ( 60.37, 19.68) {25};

\node[text=drawColor,anchor=base,inner sep=0pt, outer sep=0pt, scale=  0.88] at (125.20, 19.68) {50};

\node[text=drawColor,anchor=base,inner sep=0pt, outer sep=0pt, scale=  0.88] at (190.04, 19.68) {75};

\node[text=drawColor,anchor=base,inner sep=0pt, outer sep=0pt, scale=  0.88] at (254.87, 19.68) {100};
\end{scope}
\begin{scope}
\path[clip] (  0.00,  0.00) rectangle (325.21,216.81);
\definecolor{drawColor}{RGB}{0,0,0}

\node[text=drawColor,anchor=base,inner sep=0pt, outer sep=0pt, scale=  1.10] at (147.25,  7.64) {Age (years)};
\end{scope}
\begin{scope}
\path[clip] (  0.00,  0.00) rectangle (325.21,216.81);
\definecolor{drawColor}{RGB}{0,0,0}

\node[text=drawColor,rotate= 90.00,anchor=base,inner sep=0pt, outer sep=0pt, scale=  1.10] at ( 13.08,121.00) {Error (Percentage)};
\end{scope}
\begin{scope}
\path[clip] (  0.00,  0.00) rectangle (325.21,216.81);
\definecolor{fillColor}{RGB}{255,255,255}

\path[fill=fillColor] (273.78, 93.44) rectangle (319.71,148.56);
\end{scope}
\begin{scope}
\path[clip] (  0.00,  0.00) rectangle (325.21,216.81);
\definecolor{drawColor}{RGB}{0,0,0}

\node[text=drawColor,anchor=base west,inner sep=0pt, outer sep=0pt, scale=  1.10] at (279.28,134.41) {Gender};
\end{scope}
\begin{scope}
\path[clip] (  0.00,  0.00) rectangle (325.21,216.81);
\definecolor{fillColor}{RGB}{255,255,255}

\path[fill=fillColor] (279.28,113.39) rectangle (293.74,127.84);
\end{scope}
\begin{scope}
\path[clip] (  0.00,  0.00) rectangle (325.21,216.81);
\definecolor{drawColor}{RGB}{240,69,70}
\definecolor{fillColor}{RGB}{240,69,70}

\path[draw=drawColor,line width= 0.4pt,line join=round,line cap=round,fill=fillColor] (286.51,120.62) circle (  1.43);
\end{scope}
\begin{scope}
\path[clip] (  0.00,  0.00) rectangle (325.21,216.81);
\definecolor{fillColor}{RGB}{255,255,255}

\path[fill=fillColor] (279.28, 98.94) rectangle (293.74,113.39);
\end{scope}
\begin{scope}
\path[clip] (  0.00,  0.00) rectangle (325.21,216.81);
\definecolor{drawColor}{RGB}{98,199,107}
\definecolor{fillColor}{RGB}{98,199,107}

\path[draw=drawColor,line width= 0.4pt,line join=round,line cap=round,fill=fillColor] (286.51,106.16) circle (  1.43);
\end{scope}
\begin{scope}
\path[clip] (  0.00,  0.00) rectangle (325.21,216.81);
\definecolor{drawColor}{RGB}{0,0,0}

\node[text=drawColor,anchor=base west,inner sep=0pt, outer sep=0pt, scale=  0.88] at (299.24,117.59) {F};
\end{scope}
\begin{scope}
\path[clip] (  0.00,  0.00) rectangle (325.21,216.81);
\definecolor{drawColor}{RGB}{0,0,0}

\node[text=drawColor,anchor=base west,inner sep=0pt, outer sep=0pt, scale=  0.88] at (299.24,103.13) {M};
\end{scope}
\end{tikzpicture}

%% file: Tables/age_error_rate.tex
\begin{table}[ht]
\centering

\caption{Error rates in age prediction across FPT tools} 
\label{age_error_rate}

\begin{tabular}{lcccc}
\\[-1.8ex]\hline
\hline
Tool & No. & \multicolumn{3}{c}{Error rate - age bins}  \\
\cline{3-5}\\[-1.8ex]
& & 5 Year & 11 Year & 21 Year \\
\hline
Amazon & 32,168 & 80.56\% & 60.36\% & 34.72\% \\ 
Face++ & 31,596 & 77.85\% & 54.5\% & 25.34\% \\ 
FaceX & 31,319 & 83.48\% & 65.08\% & 42.19\% \\ 
Microsoft & 31,163 & 71.43\% & 42.2\% & 14.28\% \\ 
\hline
\end{tabular}

\end{table}

%% file: IMG/age_plot_gender_error.tex
\begin{tikzpicture}[x=1pt,y=1pt]
\definecolor{fillColor}{RGB}{255,255,255}
\path[use as bounding box,fill=fillColor,fill opacity=0.00] (0,0) rectangle (325.21,216.81);
\begin{scope}
\path[clip] (  0.00,  0.00) rectangle (325.21,216.81);
\definecolor{drawColor}{RGB}{255,255,255}
\definecolor{fillColor}{RGB}{255,255,255}

\path[draw=drawColor,line width= 0.6pt,line join=round,line cap=round,fill=fillColor] (  0.00,  0.00) rectangle (325.21,216.81);
\end{scope}
\begin{scope}
\path[clip] ( 34.64, 30.69) rectangle (262.78,211.31);
\definecolor{fillColor}{RGB}{255,255,255}

\path[fill=fillColor] ( 34.64, 30.69) rectangle (262.78,211.31);
\definecolor{drawColor}{gray}{0.92}

\path[draw=drawColor,line width= 0.3pt,line join=round] ( 34.64, 60.37) --
	(262.78, 60.37);

\path[draw=drawColor,line width= 0.3pt,line join=round] ( 34.64,115.80) --
	(262.78,115.80);

\path[draw=drawColor,line width= 0.3pt,line join=round] ( 34.64,171.23) --
	(262.78,171.23);

\path[draw=drawColor,line width= 0.3pt,line join=round] ( 94.94, 30.69) --
	( 94.94,211.31);

\path[draw=drawColor,line width= 0.3pt,line join=round] (158.96, 30.69) --
	(158.96,211.31);

\path[draw=drawColor,line width= 0.3pt,line join=round] (222.97, 30.69) --
	(222.97,211.31);

\path[draw=drawColor,line width= 0.6pt,line join=round] ( 34.64, 32.66) --
	(262.78, 32.66);

\path[draw=drawColor,line width= 0.6pt,line join=round] ( 34.64, 88.09) --
	(262.78, 88.09);

\path[draw=drawColor,line width= 0.6pt,line join=round] ( 34.64,143.52) --
	(262.78,143.52);

\path[draw=drawColor,line width= 0.6pt,line join=round] ( 34.64,198.94) --
	(262.78,198.94);

\path[draw=drawColor,line width= 0.6pt,line join=round] ( 62.94, 30.69) --
	( 62.94,211.31);

\path[draw=drawColor,line width= 0.6pt,line join=round] (126.95, 30.69) --
	(126.95,211.31);

\path[draw=drawColor,line width= 0.6pt,line join=round] (190.96, 30.69) --
	(190.96,211.31);

\path[draw=drawColor,line width= 0.6pt,line join=round] (254.97, 30.69) --
	(254.97,211.31);
\definecolor{drawColor}{RGB}{248,118,109}
\definecolor{fillColor}{RGB}{248,118,109}

\path[draw=drawColor,line width= 0.4pt,line join=round,line cap=round,fill=fillColor] ( 45.01,203.10) circle (  1.43);

\path[draw=drawColor,line width= 0.4pt,line join=round,line cap=round,fill=fillColor] ( 50.14,131.74) circle (  1.43);

\path[draw=drawColor,line width= 0.4pt,line join=round,line cap=round,fill=fillColor] ( 52.70,190.17) circle (  1.43);

\path[draw=drawColor,line width= 0.4pt,line join=round,line cap=round,fill=fillColor] ( 55.26,174.69) circle (  1.43);

\path[draw=drawColor,line width= 0.4pt,line join=round,line cap=round,fill=fillColor] ( 57.82,145.59) circle (  1.43);

\path[draw=drawColor,line width= 0.4pt,line join=round,line cap=round,fill=fillColor] ( 60.38,171.97) circle (  1.43);

\path[draw=drawColor,line width= 0.4pt,line join=round,line cap=round,fill=fillColor] ( 62.94,153.63) circle (  1.43);

\path[draw=drawColor,line width= 0.4pt,line join=round,line cap=round,fill=fillColor] ( 65.50,154.56) circle (  1.43);

\path[draw=drawColor,line width= 0.4pt,line join=round,line cap=round,fill=fillColor] ( 68.06,149.32) circle (  1.43);

\path[draw=drawColor,line width= 0.4pt,line join=round,line cap=round,fill=fillColor] ( 70.62,150.15) circle (  1.43);

\path[draw=drawColor,line width= 0.4pt,line join=round,line cap=round,fill=fillColor] ( 73.18,148.69) circle (  1.43);

\path[draw=drawColor,line width= 0.4pt,line join=round,line cap=round,fill=fillColor] ( 75.74,147.33) circle (  1.43);

\path[draw=drawColor,line width= 0.4pt,line join=round,line cap=round,fill=fillColor] ( 78.30,146.41) circle (  1.43);

\path[draw=drawColor,line width= 0.4pt,line join=round,line cap=round,fill=fillColor] ( 80.86,141.81) circle (  1.43);

\path[draw=drawColor,line width= 0.4pt,line join=round,line cap=round,fill=fillColor] ( 83.42,142.58) circle (  1.43);

\path[draw=drawColor,line width= 0.4pt,line join=round,line cap=round,fill=fillColor] ( 85.98,143.34) circle (  1.43);

\path[draw=drawColor,line width= 0.4pt,line join=round,line cap=round,fill=fillColor] ( 88.54,140.83) circle (  1.43);

\path[draw=drawColor,line width= 0.4pt,line join=round,line cap=round,fill=fillColor] ( 91.10,141.81) circle (  1.43);

\path[draw=drawColor,line width= 0.4pt,line join=round,line cap=round,fill=fillColor] ( 93.66,140.09) circle (  1.43);

\path[draw=drawColor,line width= 0.4pt,line join=round,line cap=round,fill=fillColor] ( 96.22,138.16) circle (  1.43);

\path[draw=drawColor,line width= 0.4pt,line join=round,line cap=round,fill=fillColor] ( 98.78,134.40) circle (  1.43);

\path[draw=drawColor,line width= 0.4pt,line join=round,line cap=round,fill=fillColor] (101.35,133.51) circle (  1.43);

\path[draw=drawColor,line width= 0.4pt,line join=round,line cap=round,fill=fillColor] (103.91,132.39) circle (  1.43);

\path[draw=drawColor,line width= 0.4pt,line join=round,line cap=round,fill=fillColor] (106.47,132.09) circle (  1.43);

\path[draw=drawColor,line width= 0.4pt,line join=round,line cap=round,fill=fillColor] (109.03,132.19) circle (  1.43);

\path[draw=drawColor,line width= 0.4pt,line join=round,line cap=round,fill=fillColor] (111.59,129.45) circle (  1.43);

\path[draw=drawColor,line width= 0.4pt,line join=round,line cap=round,fill=fillColor] (114.15,128.65) circle (  1.43);

\path[draw=drawColor,line width= 0.4pt,line join=round,line cap=round,fill=fillColor] (116.71,125.42) circle (  1.43);

\path[draw=drawColor,line width= 0.4pt,line join=round,line cap=round,fill=fillColor] (119.27,125.02) circle (  1.43);

\path[draw=drawColor,line width= 0.4pt,line join=round,line cap=round,fill=fillColor] (121.83,124.15) circle (  1.43);

\path[draw=drawColor,line width= 0.4pt,line join=round,line cap=round,fill=fillColor] (124.39,123.52) circle (  1.43);

\path[draw=drawColor,line width= 0.4pt,line join=round,line cap=round,fill=fillColor] (126.95,119.03) circle (  1.43);

\path[draw=drawColor,line width= 0.4pt,line join=round,line cap=round,fill=fillColor] (129.51,120.34) circle (  1.43);

\path[draw=drawColor,line width= 0.4pt,line join=round,line cap=round,fill=fillColor] (132.07,119.74) circle (  1.43);

\path[draw=drawColor,line width= 0.4pt,line join=round,line cap=round,fill=fillColor] (134.63,121.81) circle (  1.43);

\path[draw=drawColor,line width= 0.4pt,line join=round,line cap=round,fill=fillColor] (137.19,117.02) circle (  1.43);

\path[draw=drawColor,line width= 0.4pt,line join=round,line cap=round,fill=fillColor] (139.75,116.80) circle (  1.43);

\path[draw=drawColor,line width= 0.4pt,line join=round,line cap=round,fill=fillColor] (142.31,118.27) circle (  1.43);

\path[draw=drawColor,line width= 0.4pt,line join=round,line cap=round,fill=fillColor] (144.87,117.34) circle (  1.43);

\path[draw=drawColor,line width= 0.4pt,line join=round,line cap=round,fill=fillColor] (147.43,114.89) circle (  1.43);

\path[draw=drawColor,line width= 0.4pt,line join=round,line cap=round,fill=fillColor] (149.99,112.47) circle (  1.43);

\path[draw=drawColor,line width= 0.4pt,line join=round,line cap=round,fill=fillColor] (152.55,113.17) circle (  1.43);

\path[draw=drawColor,line width= 0.4pt,line join=round,line cap=round,fill=fillColor] (155.12,107.05) circle (  1.43);

\path[draw=drawColor,line width= 0.4pt,line join=round,line cap=round,fill=fillColor] (157.68,101.85) circle (  1.43);

\path[draw=drawColor,line width= 0.4pt,line join=round,line cap=round,fill=fillColor] (160.24,110.36) circle (  1.43);

\path[draw=drawColor,line width= 0.4pt,line join=round,line cap=round,fill=fillColor] (162.80,110.64) circle (  1.43);

\path[draw=drawColor,line width= 0.4pt,line join=round,line cap=round,fill=fillColor] (165.36,110.80) circle (  1.43);

\path[draw=drawColor,line width= 0.4pt,line join=round,line cap=round,fill=fillColor] (167.92,104.68) circle (  1.43);

\path[draw=drawColor,line width= 0.4pt,line join=round,line cap=round,fill=fillColor] (170.48,101.22) circle (  1.43);

\path[draw=drawColor,line width= 0.4pt,line join=round,line cap=round,fill=fillColor] (173.04, 96.18) circle (  1.43);

\path[draw=drawColor,line width= 0.4pt,line join=round,line cap=round,fill=fillColor] (175.60,110.78) circle (  1.43);

\path[draw=drawColor,line width= 0.4pt,line join=round,line cap=round,fill=fillColor] (178.16, 96.14) circle (  1.43);

\path[draw=drawColor,line width= 0.4pt,line join=round,line cap=round,fill=fillColor] (180.72, 88.09) circle (  1.43);

\path[draw=drawColor,line width= 0.4pt,line join=round,line cap=round,fill=fillColor] (183.28, 95.13) circle (  1.43);

\path[draw=drawColor,line width= 0.4pt,line join=round,line cap=round,fill=fillColor] (185.84, 80.64) circle (  1.43);

\path[draw=drawColor,line width= 0.4pt,line join=round,line cap=round,fill=fillColor] (188.40, 89.94) circle (  1.43);

\path[draw=drawColor,line width= 0.4pt,line join=round,line cap=round,fill=fillColor] (190.96, 94.15) circle (  1.43);

\path[draw=drawColor,line width= 0.4pt,line join=round,line cap=round,fill=fillColor] (193.52,126.54) circle (  1.43);

\path[draw=drawColor,line width= 0.4pt,line join=round,line cap=round,fill=fillColor] (206.32,117.88) circle (  1.43);
\definecolor{drawColor}{RGB}{0,186,56}
\definecolor{fillColor}{RGB}{0,186,56}

\path[draw=drawColor,line width= 0.4pt,line join=round,line cap=round,fill=fillColor] ( 52.70,182.18) circle (  1.43);

\path[draw=drawColor,line width= 0.4pt,line join=round,line cap=round,fill=fillColor] ( 55.26,163.35) circle (  1.43);

\path[draw=drawColor,line width= 0.4pt,line join=round,line cap=round,fill=fillColor] ( 57.82,167.17) circle (  1.43);

\path[draw=drawColor,line width= 0.4pt,line join=round,line cap=round,fill=fillColor] ( 60.38,164.13) circle (  1.43);

\path[draw=drawColor,line width= 0.4pt,line join=round,line cap=round,fill=fillColor] ( 62.94,165.83) circle (  1.43);

\path[draw=drawColor,line width= 0.4pt,line join=round,line cap=round,fill=fillColor] ( 65.50,160.99) circle (  1.43);

\path[draw=drawColor,line width= 0.4pt,line join=round,line cap=round,fill=fillColor] ( 68.06,160.26) circle (  1.43);

\path[draw=drawColor,line width= 0.4pt,line join=round,line cap=round,fill=fillColor] ( 70.62,158.71) circle (  1.43);

\path[draw=drawColor,line width= 0.4pt,line join=round,line cap=round,fill=fillColor] ( 73.18,157.20) circle (  1.43);

\path[draw=drawColor,line width= 0.4pt,line join=round,line cap=round,fill=fillColor] ( 75.74,156.08) circle (  1.43);

\path[draw=drawColor,line width= 0.4pt,line join=round,line cap=round,fill=fillColor] ( 78.30,155.38) circle (  1.43);

\path[draw=drawColor,line width= 0.4pt,line join=round,line cap=round,fill=fillColor] ( 80.86,153.87) circle (  1.43);

\path[draw=drawColor,line width= 0.4pt,line join=round,line cap=round,fill=fillColor] ( 83.42,152.66) circle (  1.43);

\path[draw=drawColor,line width= 0.4pt,line join=round,line cap=round,fill=fillColor] ( 85.98,152.72) circle (  1.43);

\path[draw=drawColor,line width= 0.4pt,line join=round,line cap=round,fill=fillColor] ( 88.54,152.14) circle (  1.43);

\path[draw=drawColor,line width= 0.4pt,line join=round,line cap=round,fill=fillColor] ( 91.10,150.33) circle (  1.43);

\path[draw=drawColor,line width= 0.4pt,line join=round,line cap=round,fill=fillColor] ( 93.66,148.50) circle (  1.43);

\path[draw=drawColor,line width= 0.4pt,line join=round,line cap=round,fill=fillColor] ( 96.22,149.27) circle (  1.43);

\path[draw=drawColor,line width= 0.4pt,line join=round,line cap=round,fill=fillColor] ( 98.78,147.48) circle (  1.43);

\path[draw=drawColor,line width= 0.4pt,line join=round,line cap=round,fill=fillColor] (101.35,145.67) circle (  1.43);

\path[draw=drawColor,line width= 0.4pt,line join=round,line cap=round,fill=fillColor] (103.91,145.30) circle (  1.43);

\path[draw=drawColor,line width= 0.4pt,line join=round,line cap=round,fill=fillColor] (106.47,143.87) circle (  1.43);

\path[draw=drawColor,line width= 0.4pt,line join=round,line cap=round,fill=fillColor] (109.03,142.84) circle (  1.43);

\path[draw=drawColor,line width= 0.4pt,line join=round,line cap=round,fill=fillColor] (111.59,141.75) circle (  1.43);

\path[draw=drawColor,line width= 0.4pt,line join=round,line cap=round,fill=fillColor] (114.15,140.57) circle (  1.43);

\path[draw=drawColor,line width= 0.4pt,line join=round,line cap=round,fill=fillColor] (116.71,139.02) circle (  1.43);

\path[draw=drawColor,line width= 0.4pt,line join=round,line cap=round,fill=fillColor] (119.27,138.95) circle (  1.43);

\path[draw=drawColor,line width= 0.4pt,line join=round,line cap=round,fill=fillColor] (121.83,136.86) circle (  1.43);

\path[draw=drawColor,line width= 0.4pt,line join=round,line cap=round,fill=fillColor] (124.39,136.08) circle (  1.43);

\path[draw=drawColor,line width= 0.4pt,line join=round,line cap=round,fill=fillColor] (126.95,135.31) circle (  1.43);

\path[draw=drawColor,line width= 0.4pt,line join=round,line cap=round,fill=fillColor] (129.51,133.18) circle (  1.43);

\path[draw=drawColor,line width= 0.4pt,line join=round,line cap=round,fill=fillColor] (132.07,133.15) circle (  1.43);

\path[draw=drawColor,line width= 0.4pt,line join=round,line cap=round,fill=fillColor] (134.63,130.41) circle (  1.43);

\path[draw=drawColor,line width= 0.4pt,line join=round,line cap=round,fill=fillColor] (137.19,129.73) circle (  1.43);

\path[draw=drawColor,line width= 0.4pt,line join=round,line cap=round,fill=fillColor] (139.75,127.50) circle (  1.43);

\path[draw=drawColor,line width= 0.4pt,line join=round,line cap=round,fill=fillColor] (142.31,126.66) circle (  1.43);

\path[draw=drawColor,line width= 0.4pt,line join=round,line cap=round,fill=fillColor] (144.87,126.16) circle (  1.43);

\path[draw=drawColor,line width= 0.4pt,line join=round,line cap=round,fill=fillColor] (147.43,124.71) circle (  1.43);

\path[draw=drawColor,line width= 0.4pt,line join=round,line cap=round,fill=fillColor] (149.99,121.79) circle (  1.43);

\path[draw=drawColor,line width= 0.4pt,line join=round,line cap=round,fill=fillColor] (152.55,121.24) circle (  1.43);

\path[draw=drawColor,line width= 0.4pt,line join=round,line cap=round,fill=fillColor] (155.12,120.81) circle (  1.43);

\path[draw=drawColor,line width= 0.4pt,line join=round,line cap=round,fill=fillColor] (157.68,120.69) circle (  1.43);

\path[draw=drawColor,line width= 0.4pt,line join=round,line cap=round,fill=fillColor] (160.24,117.52) circle (  1.43);

\path[draw=drawColor,line width= 0.4pt,line join=round,line cap=round,fill=fillColor] (162.80,117.19) circle (  1.43);

\path[draw=drawColor,line width= 0.4pt,line join=round,line cap=round,fill=fillColor] (165.36,118.84) circle (  1.43);

\path[draw=drawColor,line width= 0.4pt,line join=round,line cap=round,fill=fillColor] (167.92,115.20) circle (  1.43);

\path[draw=drawColor,line width= 0.4pt,line join=round,line cap=round,fill=fillColor] (170.48,114.51) circle (  1.43);

\path[draw=drawColor,line width= 0.4pt,line join=round,line cap=round,fill=fillColor] (173.04,111.15) circle (  1.43);

\path[draw=drawColor,line width= 0.4pt,line join=round,line cap=round,fill=fillColor] (175.60,111.19) circle (  1.43);

\path[draw=drawColor,line width= 0.4pt,line join=round,line cap=round,fill=fillColor] (178.16,112.17) circle (  1.43);

\path[draw=drawColor,line width= 0.4pt,line join=round,line cap=round,fill=fillColor] (180.72,110.42) circle (  1.43);

\path[draw=drawColor,line width= 0.4pt,line join=round,line cap=round,fill=fillColor] (183.28,110.78) circle (  1.43);

\path[draw=drawColor,line width= 0.4pt,line join=round,line cap=round,fill=fillColor] (185.84,107.33) circle (  1.43);

\path[draw=drawColor,line width= 0.4pt,line join=round,line cap=round,fill=fillColor] (188.40,112.24) circle (  1.43);

\path[draw=drawColor,line width= 0.4pt,line join=round,line cap=round,fill=fillColor] (190.96,100.94) circle (  1.43);

\path[draw=drawColor,line width= 0.4pt,line join=round,line cap=round,fill=fillColor] (193.52,102.59) circle (  1.43);

\path[draw=drawColor,line width= 0.4pt,line join=round,line cap=round,fill=fillColor] (196.08,107.14) circle (  1.43);

\path[draw=drawColor,line width= 0.4pt,line join=round,line cap=round,fill=fillColor] (198.64,101.78) circle (  1.43);

\path[draw=drawColor,line width= 0.4pt,line join=round,line cap=round,fill=fillColor] (201.20, 98.02) circle (  1.43);

\path[draw=drawColor,line width= 0.4pt,line join=round,line cap=round,fill=fillColor] (203.76, 95.24) circle (  1.43);

\path[draw=drawColor,line width= 0.4pt,line join=round,line cap=round,fill=fillColor] (206.32, 91.46) circle (  1.43);

\path[draw=drawColor,line width= 0.4pt,line join=round,line cap=round,fill=fillColor] (208.89, 91.89) circle (  1.43);

\path[draw=drawColor,line width= 0.4pt,line join=round,line cap=round,fill=fillColor] (211.45,103.58) circle (  1.43);

\path[draw=drawColor,line width= 0.4pt,line join=round,line cap=round,fill=fillColor] (214.01, 95.71) circle (  1.43);

\path[draw=drawColor,line width= 0.4pt,line join=round,line cap=round,fill=fillColor] (216.57,117.19) circle (  1.43);

\path[draw=drawColor,line width= 0.4pt,line join=round,line cap=round,fill=fillColor] (219.13, 82.55) circle (  1.43);

\path[draw=drawColor,line width= 0.4pt,line join=round,line cap=round,fill=fillColor] (221.69, 60.37) circle (  1.43);

\path[draw=drawColor,line width= 0.4pt,line join=round,line cap=round,fill=fillColor] (229.37, 66.61) circle (  1.43);

\path[draw=drawColor,line width= 0.4pt,line join=round,line cap=round,fill=fillColor] (231.93, 38.90) circle (  1.43);

\path[draw=drawColor,line width= 0.4pt,line join=round,line cap=round,fill=fillColor] (237.05, 77.93) circle (  1.43);

\path[draw=drawColor,line width= 0.4pt,line join=round,line cap=round,fill=fillColor] (242.17, 83.24) circle (  1.43);

\path[draw=drawColor,line width= 0.4pt,line join=round,line cap=round,fill=fillColor] (247.29, 81.85) circle (  1.43);

\path[draw=drawColor,line width= 0.4pt,line join=round,line cap=round,fill=fillColor] (252.41, 77.00) circle (  1.43);
\definecolor{drawColor}{RGB}{97,156,255}
\definecolor{fillColor}{RGB}{97,156,255}

\path[draw=drawColor,line width= 0.4pt,line join=round,line cap=round,fill=fillColor] ( 62.94,155.29) circle (  1.43);

\path[draw=drawColor,line width= 0.4pt,line join=round,line cap=round,fill=fillColor] ( 65.50,174.00) circle (  1.43);

\path[draw=drawColor,line width= 0.4pt,line join=round,line cap=round,fill=fillColor] ( 68.06,151.83) circle (  1.43);

\path[draw=drawColor,line width= 0.4pt,line join=round,line cap=round,fill=fillColor] ( 70.62,142.13) circle (  1.43);

\path[draw=drawColor,line width= 0.4pt,line join=round,line cap=round,fill=fillColor] ( 83.42,166.15) circle (  1.43);

\path[draw=drawColor,line width= 0.4pt,line join=round,line cap=round,fill=fillColor] ( 85.98,165.69) circle (  1.43);

\path[draw=drawColor,line width= 0.4pt,line join=round,line cap=round,fill=fillColor] ( 88.54,144.21) circle (  1.43);

\path[draw=drawColor,line width= 0.4pt,line join=round,line cap=round,fill=fillColor] ( 93.66,160.14) circle (  1.43);

\path[draw=drawColor,line width= 0.4pt,line join=round,line cap=round,fill=fillColor] (106.47,108.87) circle (  1.43);

\path[draw=drawColor,line width= 0.4pt,line join=round,line cap=round,fill=fillColor] (116.71,137.28) circle (  1.43);

\path[draw=drawColor,line width= 0.4pt,line join=round,line cap=round,fill=fillColor] (134.63,147.67) circle (  1.43);

\path[draw=drawColor,line width= 0.4pt,line join=round,line cap=round,fill=fillColor] (160.24, 79.77) circle (  1.43);
\definecolor{drawColor}{RGB}{0,0,0}

\path[draw=drawColor,line width= 0.6pt,line join=round] ( 34.64,143.52) -- (262.78,143.52);
\definecolor{drawColor}{gray}{0.20}

\path[draw=drawColor,line width= 0.6pt,line join=round,line cap=round] ( 34.64, 30.69) rectangle (262.78,211.31);
\end{scope}
\begin{scope}
\path[clip] (  0.00,  0.00) rectangle (325.21,216.81);
\definecolor{drawColor}{gray}{0.30}

\node[text=drawColor,anchor=base east,inner sep=0pt, outer sep=0pt, scale=  0.88] at ( 29.69, 29.63) {-40};

\node[text=drawColor,anchor=base east,inner sep=0pt, outer sep=0pt, scale=  0.88] at ( 29.69, 85.06) {-20};

\node[text=drawColor,anchor=base east,inner sep=0pt, outer sep=0pt, scale=  0.88] at ( 29.69,140.48) {0};

\node[text=drawColor,anchor=base east,inner sep=0pt, outer sep=0pt, scale=  0.88] at ( 29.69,195.91) {20};
\end{scope}
\begin{scope}
\path[clip] (  0.00,  0.00) rectangle (325.21,216.81);
\definecolor{drawColor}{gray}{0.20}

\path[draw=drawColor,line width= 0.6pt,line join=round] ( 31.89, 32.66) --
	( 34.64, 32.66);

\path[draw=drawColor,line width= 0.6pt,line join=round] ( 31.89, 88.09) --
	( 34.64, 88.09);

\path[draw=drawColor,line width= 0.6pt,line join=round] ( 31.89,143.52) --
	( 34.64,143.52);

\path[draw=drawColor,line width= 0.6pt,line join=round] ( 31.89,198.94) --
	( 34.64,198.94);
\end{scope}
\begin{scope}
\path[clip] (  0.00,  0.00) rectangle (325.21,216.81);
\definecolor{drawColor}{gray}{0.20}

\path[draw=drawColor,line width= 0.6pt,line join=round] ( 62.94, 27.94) --
	( 62.94, 30.69);

\path[draw=drawColor,line width= 0.6pt,line join=round] (126.95, 27.94) --
	(126.95, 30.69);

\path[draw=drawColor,line width= 0.6pt,line join=round] (190.96, 27.94) --
	(190.96, 30.69);

\path[draw=drawColor,line width= 0.6pt,line join=round] (254.97, 27.94) --
	(254.97, 30.69);
\end{scope}
\begin{scope}
\path[clip] (  0.00,  0.00) rectangle (325.21,216.81);
\definecolor{drawColor}{gray}{0.30}

\node[text=drawColor,anchor=base,inner sep=0pt, outer sep=0pt, scale=  0.88] at ( 62.94, 19.68) {25};

\node[text=drawColor,anchor=base,inner sep=0pt, outer sep=0pt, scale=  0.88] at (126.95, 19.68) {50};

\node[text=drawColor,anchor=base,inner sep=0pt, outer sep=0pt, scale=  0.88] at (190.96, 19.68) {75};

\node[text=drawColor,anchor=base,inner sep=0pt, outer sep=0pt, scale=  0.88] at (254.97, 19.68) {100};
\end{scope}
\begin{scope}
\path[clip] (  0.00,  0.00) rectangle (325.21,216.81);
\definecolor{drawColor}{RGB}{0,0,0}

\node[text=drawColor,anchor=base,inner sep=0pt, outer sep=0pt, scale=  1.10] at (148.71,  7.64) {Age (years)};
\end{scope}
\begin{scope}
\path[clip] (  0.00,  0.00) rectangle (325.21,216.81);
\definecolor{drawColor}{RGB}{0,0,0}

\node[text=drawColor,rotate= 90.00,anchor=base,inner sep=0pt, outer sep=0pt, scale=  1.10] at ( 13.08,121.00) {Predicted age -- actual age (years)};
\end{scope}
\begin{scope}
\path[clip] (  0.00,  0.00) rectangle (325.21,216.81);
\definecolor{fillColor}{RGB}{255,255,255}

\path[fill=fillColor] (273.78, 86.21) rectangle (319.71,155.79);
\end{scope}
\begin{scope}
\path[clip] (  0.00,  0.00) rectangle (325.21,216.81);
\definecolor{drawColor}{RGB}{0,0,0}

\node[text=drawColor,anchor=base west,inner sep=0pt, outer sep=0pt, scale=  1.10] at (279.28,141.64) {Gender};
\end{scope}
\begin{scope}
\path[clip] (  0.00,  0.00) rectangle (325.21,216.81);
\definecolor{fillColor}{RGB}{255,255,255}

\path[fill=fillColor] (279.28,120.62) rectangle (293.74,135.07);
\end{scope}
\begin{scope}
\path[clip] (  0.00,  0.00) rectangle (325.21,216.81);
\definecolor{drawColor}{RGB}{248,118,109}
\definecolor{fillColor}{RGB}{248,118,109}

\path[draw=drawColor,line width= 0.4pt,line join=round,line cap=round,fill=fillColor] (286.51,127.84) circle (  1.43);
\end{scope}
\begin{scope}
\path[clip] (  0.00,  0.00) rectangle (325.21,216.81);
\definecolor{fillColor}{RGB}{255,255,255}

\path[fill=fillColor] (279.28,106.16) rectangle (293.74,120.62);
\end{scope}
\begin{scope}
\path[clip] (  0.00,  0.00) rectangle (325.21,216.81);
\definecolor{drawColor}{RGB}{0,186,56}
\definecolor{fillColor}{RGB}{0,186,56}

\path[draw=drawColor,line width= 0.4pt,line join=round,line cap=round,fill=fillColor] (286.51,113.39) circle (  1.43);
\end{scope}
\begin{scope}
\path[clip] (  0.00,  0.00) rectangle (325.21,216.81);
\definecolor{fillColor}{RGB}{255,255,255}

\path[fill=fillColor] (279.28, 91.71) rectangle (293.74,106.16);
\end{scope}
\begin{scope}
\path[clip] (  0.00,  0.00) rectangle (325.21,216.81);
\definecolor{drawColor}{RGB}{97,156,255}
\definecolor{fillColor}{RGB}{97,156,255}

\path[draw=drawColor,line width= 0.4pt,line join=round,line cap=round,fill=fillColor] (286.51, 98.94) circle (  1.43);
\end{scope}
\begin{scope}
\path[clip] (  0.00,  0.00) rectangle (325.21,216.81);
\definecolor{drawColor}{RGB}{0,0,0}

\node[text=drawColor,anchor=base west,inner sep=0pt, outer sep=0pt, scale=  0.88] at (299.24,124.81) {F};
\end{scope}
\begin{scope}
\path[clip] (  0.00,  0.00) rectangle (325.21,216.81);
\definecolor{drawColor}{RGB}{0,0,0}

\node[text=drawColor,anchor=base west,inner sep=0pt, outer sep=0pt, scale=  0.88] at (299.24,110.36) {M};
\end{scope}
\begin{scope}
\path[clip] (  0.00,  0.00) rectangle (325.21,216.81);
\definecolor{drawColor}{RGB}{0,0,0}

\node[text=drawColor,anchor=base west,inner sep=0pt, outer sep=0pt, scale=  0.88] at (299.24, 95.91) {T};
\end{scope}
\end{tikzpicture}

%% file: 06_conclusion.tex
\section{Conclusion} \label{sec6}

In this paper we audited the performance of four commercial facial processing tools, Amazon's Rekognition, Microsoft Azure's Face, Face++ and FaceX, on a novel India faces dataset gathered from the \gls{eci}'s election candidates portal. The audit focused on the functions of face detection, gender classification and age prediction and did not include face recognition. To the best of our knowledge, this is the first study of this nature that carries out an empirical evaluation of the performance of facial processing tools on Indian faces.

Our findings point to significant variations in the performance of the tools. There were also some clear group-specific trends observed across all tools. For face detection, Microsoft reported the highest detection error rate of 3.17\%, implying an inability to correctly detect over a thousand faces in the dataset. This figure was similarly high for the Indian company, FaceX. Non-detection means that these images cannot be subjected to any further types of facial processing. While this may be a favourable outcome in certain cases, such as detection by surveillance systems, it can also lead to exclusions and hardships in the context of mandatory or hard to avoid facial processing systems.

All the tools reported higher errors in the gender classification of females compared to males and a complete inability to correctly classify persons belonging to the third gender. The female classification error was the highest in the case of China-based Face++. It had an error rate of almost 15\% for Indian female faces. Besides being significantly higher than Face++'s error rate for Indian males, which was almost zero, this error rate is also much higher than that the classification errors found for females of other nationalities in a previous study. An audit conducted in 2018 reported an error rate of 2.5\% for Face++ using an image dataset of African and European faces \autocite{rajiBuolamwini2019}. This points to the lack of transferability of findings from audits done using datasets of non-Indian faces into the Indian context. It also illustrates the tendency of commercial firms to introduce band-aid measures designed to address specific issues that may be highlighted by researchers and civil society.

Age prediction errors were also high across all tools. Despite taking into account an acceptable error margin of plus or minus 10 years from a person's actual age, the tools displayed age prediction failures in the range of 14.3\% to 42.2\%. There was also a clear trend of predicting lower age ranges -- as one grows older the tools are more likely to predict a younger age.

Future work on this subject can extend in at least two directions. The first would be to look at how facial processing interacts with other axes of diversity in the Indian society, such as caste, tribe, religion, and skin tones, and the intersection of multiple factors. The second would be to expand the scope of review beyond face detection and analysis to also include face recognition systems that carry out verification and identification functions. However, the ability to do so will have to be shaped by the limitations of available datasets and important ethical considerations.

Finally, we would like to emphasise that the accuracy of facial processing is only one among several necessary elements of a fair and accountable system. In fact, accuracy could itself be a double-edged sword, with higher accuracy leading to more rampant adoption of the technology and more exploitative use cases. Therefore, along with the need for more context-specific research directed at auditing the efficacy of facial processing systems, we must also persist with the broader agenda of questioning the validity and adoption of such systems, particularly in light of India's legal and institutional realities.